\documentclass[11pt,a4paper]{article}
\usepackage[hyperref]{acl2020}
\usepackage{url}  
\usepackage{times}  
\usepackage{helvet}  
\usepackage{courier}  
\usepackage{url}  
\usepackage{graphicx}  
\usepackage{algorithmic}
\usepackage{amsmath}
\usepackage{hyperref}
\usepackage{balance}
\usepackage{booktabs} 
\usepackage{paralist}
\usepackage{verbatim}
\usepackage{tablefootnote}
\makeatletter
\g@addto@macro{\UrlBreaks}{\UrlOrds}
\makeatother
\aclfinalcopy

\title{Hater-O-Genius Aggression Classification using Capsule Networks}

\author{
Parth Patwa\textsuperscript{1} \quad 
Srinivas PYKL\textsuperscript{1} \quad
Amitava Das\textsuperscript{2}\quad \\
\bf Prerana Mukherjee\textsuperscript{1}  \quad
\bf Viswanath Pulabaigari\textsuperscript{1} \quad \\
\textsuperscript{1}Indian Institute of Information Technology Sri City, India \quad
\textsuperscript{2}Wipro AI Labs, India\\
\tt \textsuperscript{1}\{parthprasad.p17, srinivas.p, prerana.m, viswanath.p\}@iiits.in\\
\tt\textsuperscript{2}amitava.das2@wipro.com
}

\begin{document}
\maketitle
\begin{abstract}
Contending hate speech in social media is one of the most challenging social problems of our time. There are various types of anti-social behavior in social media. Foremost of them is aggressive behavior, which is causing many social issues such as affecting the social lives and mental health of social media users. In this paper, we propose an end-to-end ensemble-based architecture to automatically identify and classify aggressive tweets. Tweets are classified into three categories - Covertly Aggressive, Overtly Aggressive, and Non-Aggressive. 
The proposed architecture is an ensemble of smaller subnetworks that are able to characterize the feature embeddings effectively. We demonstrate qualitatively that each of the smaller subnetworks is able to learn unique features. Our best model is an ensemble of Capsule Networks and results in a 65.2\% F1 score on the Facebook test set, which results in a performance gain of 0.95\% over the TRAC-2018 winners. The code and the model weights are publicly available at \url{https://github.com/parthpatwa/Hater-O-Genius-Aggression-Classification-using-Capsule-Networks}.

\end{abstract}

\section{Introduction}
Even though social media offers several benefits to people, it has caused some negative effects due to the misuse of freedom of speech by a few people.  \par 
Aggression is a behavior that is intended to harm other individuals who do not wish to be harmed \cite{doi:10.1002/1098-2337(1994)20:6<461::AID-AB2480200606>3.0.CO;2-O}. Aggressive words are commonly used to inflict mental pain on the victim by showing covert aggression, overt aggression or by using offensive language \cite{hateoffensive}. 

The process of manually weeding out aggressive tweets from social media is expensive and indefinitely slow. So, there is a growing need to build and analyze automatic aggression classifiers. \nocite{10.1007/978-3-642-36973-5_62}
\nocite{DBLP:journals/corr/abs-1803-09402}
\par
In this paper, we propose an architecture that is an ensemble of multiple subnetworks to identify aggressive tweets, where each subnetwork learns unique features. We explore different word embeddings for dense representation \cite{wordrepresentation}, deep learning (CNN, LSTM), and Capsule Networks \cite{2017arXiv171009829S}. Our best model (figure \ref{fig:capsulnet}) uses Capsule Network, and gives a 65.20\% F1 score, which is a 0.95\% improvement over the model proposed by \citet{Aroyehun:trac18}. We also release the code and the model weights.

\section{Related Work}
The challenge of tackling antisocial behavior like abuse, hate speech, and aggression on social media has recently received much attention. Researchers like \citet{Nobata:2016:ALD:2872427.2883062} tried detecting abusive language by using Machine Learning and linguistic features. Other researchers like \citet{hatespeech} used CNNs and LSTMs, along with gradient boosting, to detect hate speech. 

The TRAC-2018 shared task \cite{tracsharedtask}, aimed to detect aggression, was won by \citet{Aroyehun:trac18}, who used deep learning, data augmentation, and pseudo labeling to get a 64.25\% F1 score. Another team \citet{Risch:trac18}, used deep learning along with data augmentation and hand-picked features to detect aggression. However, in order to develop an end-to-end automated system, one cannot use hand-picked features as they may vary from dataset to dataset. \citet{Srivastava:trac18} experimented with capsulenets for detecting aggression and achieved a 63.43\% F1 score. Our work differs from theirs as we experiment with architectures (Fig. \ref{fig:capsulnet}) that are an ensemble of multiple subnetworks. Recently, \citet{Kahndelwal-sota} used pooled biLSTM and NLP features to achieve 67.7\% F1 score on the TRAC-2018 Facebook data. 

The TRAC-2020 shared task \cite{kumar-etal-2020-evaluating} released a data set \cite{Bhattacharja:10} of aggression and misogyny in Hindi, English and Bengali posts. \citet{risch-krestel-2020-bagging} tried an ensemble of BERT to achieve the best performance on most tasks. \citet{safi-samghabadi-etal-2020-aggression} used BERT in a multi-task manner to solve the task, whereas \citet{kumari-singh-2020-ai} used LSTM and CNNs. 

\section{Dataset}
To identify the type of aggression, we use the English train dataset, and the Facebook (fb) test dataset provided by the 2018 TRAC shared task \cite{tracsharedtask}. The data collection and annotation method is described in \citet{DBLP:journals/corr/abs-1803-09402}. The training data is combined with the augmented data provided by \citet{Risch:trac18}. The final distribution is given in table \ref{tab:Aggressive}. The data has English-Hindi code-mixed tweets, which are annotated with one of three labels:

\begin{itemize}
    \item \textbf{Covertly Aggressive (CAG)}: Behavior that seeks to indirectly harm the victim by using satire and sarcasm \cite{DBLP:journals/corr/abs-1803-09402}. E.g., \textit{"Irony is your display picture at one end you are happy seeing some one innocent dying and at other end you are praying to not kill an innocent"}
    
    \item \textbf{Overtly Aggressive (OAG)}: Direct and explicit form of aggression which includes derogatory comparison, verbal attack or abusive words towards a group or an individual \cite{W18-4408}. E.g., \textit{"Shame on you assholes showing some other video and making it a fake news u chooths i hope each one you at *** news will  rot in hell"}

    \item \textbf{NAG}: Texts which are not aggressive. E.g., \textit{"hope car occupants are safe and unharmed."}
    
\end{itemize}
\begin{table}[t!]
\centering
\caption{Data distribution}
\label{tab:Aggressive}
\begin{tabular}{l|l|r}
    \toprule
Class & Train &Test  \\ 
    \midrule
Covertly  Aggressive & 14,187 & 144    \\
Overtly Aggressive       & 9,137 & 142   \\ 
Non-Aggressive      & 16,188 & 630   \\ 
    \midrule
Total       & 39,512 &916   \\ 
\bottomrule
\end{tabular}
\end{table}
We observe that the dataset contains some tweets which have improbable annotations. For example, the tweet \textit{"Mr. Sun you are wrong, Pakistan produces one thing that is ' terrorists' and through CPEC Pak will increase the supply of this product throughout world. Wait you will feel the touch of their product in your Muslim dominated province."} is labeled as NAG; \textit{"\#salute you my friend"} is labeled as OAG. To have a fair comparison with the results of previous works, we don't do anything to address this. The dataset is imbalanced with maximum tweets labeled as NAG.

\section{Preprocessing and Embeddings}
\begin{figure*}[t!]
\centering
\includegraphics[scale =.5]{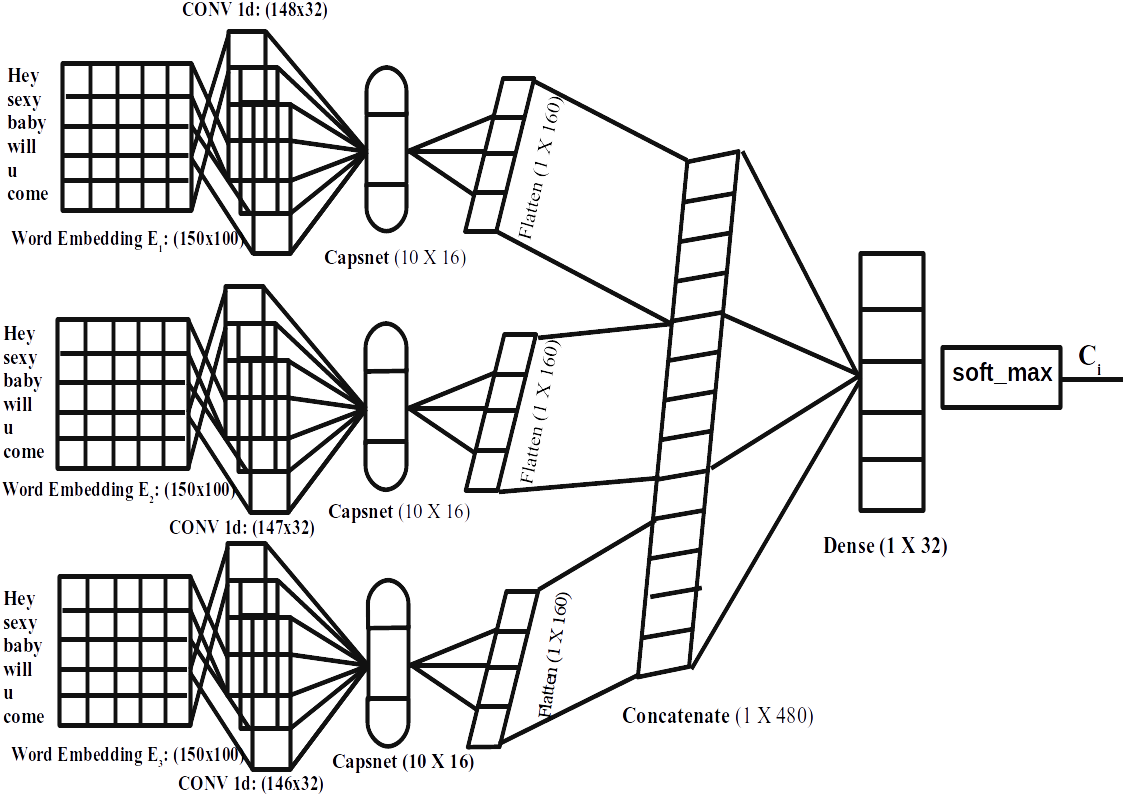}
\caption{Architecture of CN1 model}
 \label{fig:capsulnet}
\end{figure*}

The tweets are first converted to lower case. Next, we remove digits, special characters, emojis, urls, and stop words. We restrict the continuous repetition of the same character in a word to 2 (e.g. 'suuuuuuper' is converted to 'suuper'). Each tweet is tokenized and converted into a sequence of integers. The maximum sequence length is restricted to 150. To have dense representation of tokens, the following word embedding features are used:

\begin{itemize}
\item\textbf{Glove++:}{ Given the word, we first check whether it is present in
Glove pre-trained 6b 100d embeddings, and use the embedding if it exists. For Out-Of-Vocabulary words, we use the word vectors that we train on the entire data using the Gensim library.
 }

\item\textbf{Aggression Embeddings: }{To have distinguishing features to separate aggressive tweets from non-aggressive tweets, we create aggression word embeddings. We take all the tweets classified as OAG and CAG and train word vectors on them.} 

\item\textbf{Char Trigram:} {To get sub-word information, we create character trigram embeddings.}
\end{itemize}

\section{Proposed Architecture}
We propose an architecture that combines features that are learned from an ensemble of subnetworks and leverages the feature representation to classify aggression. All models optimize the categorical crossentropy loss function using adam optimizer. All the dense layers, except the final layer, have ReLu activation. All the CNN layers are followed by dropout = 0.5. Every model is an ensemble of smaller subnetworks. Each subnetwork (SN) has the following configuration:

\begin{itemize}
    \item \textbf{Embedding layer} - Each token in the input sequence is represented by its word vector. Word embeddings help to capture the meaning of the word.
  
    \item \textbf{Convolutional layer} - A convolutional layer, having reLu activation function, to extract spatial features.
   
    \item \textbf{Max-pooling layer} of size 2 or 3 in case of Deep Learning models.
    
    \item \textbf{Capsule layer} to better preserve spatial information, in case of Capsulenet models.
\end{itemize}

Each SN of the model uses a different configuration for the CNN layer or embedding. Therefore each SN learns different information and generates different features. The output of each SN is flattened and merged and is passed as input to dense layers. The last dense layer has three neurons and a softmax activation function, which gives a probability score to each of the three classes, and the one with the highest score is the predicted class. 
\subsection{Deep Learning (DL) Models}
The following are the DL baselines: 

\textbf{DL1:} It is an ensemble of three subnetworks. All three SNs use Glove++ embeddings for the embedding layer. The CNN layers in each SN have kernel sizes 3,5 and 7, respectively.\newline
\textbf{DL2:} It is an ensemble of 9 SNs.  Each max-pooling layer is followed by a biLSTM layer, having 200 units, to capture long term dependencies. SN 1-3 use Glove++ embeddings. SN 4-6 use Aggression embeddings. SN 7-9 use Character-level trigram embeddings. CNN layer in SN 1,4,7 has kernel size = 3, in SN 2,5,8 has kernel size = 5 and in SN 3,6,9 has kernel size = 7. 
\subsection{Capsule Network (CN) Models }
The main difference between CN models and DL models is that the CN models use a capsule layer instead of max-pooling layer. The capsule layer has 10 capsules of 16 dimension each. Max-pooling reduces computational complexity but leads to the loss of spatial information.

Capsules are a group of neurons that are represented as vectors. The orientation of the feature vector is preserved in capsules. They use a function called squashing for non-linearity. Dynamic Routing is used to route the feature vector of the lower-level capsule to the appropriate next level capsule \cite{2017arXiv171009829S}. Dynamic Routing is based on a coupling coefficient that measures the similarity between vectors that predict the upper capsule and the lower capsule and learns which lower capsule should be directed to which upper capsule \cite{DBLP:journals/corr/abs-1808-03976}. Through this process, capsule layers preserve spatial information, learn semantic representation, and ignore words that are insignificant.  \newline
\textbf{CN1: }The architecture is shown in figure \ref{fig:capsulnet}. It is an ensemble of 3 subnetworks. Each SN uses Glove++ embeddings, and the CNN layers have kernel size = 3,4 and 5, respectively.\newline
\textbf{CN2: }Like CN1, but there is an additional biLSTM layer, having 300 units, after the capsule layer.

\section{Results and Discussion}
\begin{table}
\centering
\begin{tabular}{|l|l|l|l|}
\hline
\multicolumn{2}{|l|}{DL models} & \multicolumn{2}{l|}{CN models}  \\ \hline
DL1 & 57.17\% & CN1 & \textbf{65.20}\%   \\ \hline
DL2 & 60.34\% & CN2 & 62.70\%   \\ \hline  
\end{tabular}
\caption{Weighted F1 scores of DL and CN models}
\label{tab:results}
\end{table}
From table \ref{tab:results}, we see that the CN models perform better than DL models. Both the CN models are comparable to the models proposed by \citet{Srivastava:trac18}. This validates the usefulness of capsule networks for aggression detection. CN1 gives the best results and is better than the best model proposed by \citet{Aroyehun:trac18}. DL2 works better than DL1, as it captures more information. The performance drops from CN1 to CN2, despite CN2 having an additional biLSTM layer. This shows that a more complex model is not necessarily better, which is in agreement with the observations of \citet{Aroyehun:trac18}. This could be due to over-fitting.
\begin{figure}[t!]
\centering
\includegraphics[scale = .5]{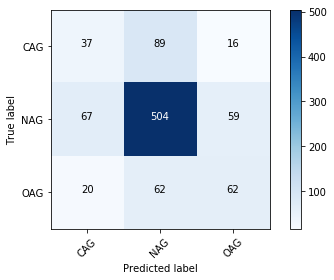} 
\captionof{figure}{ Confusion matrix of CN1 model}
 \label{fig:confusion matrix}
\end{figure}
\begin{figure}[t!]
\centering
\includegraphics[width = .6\linewidth]{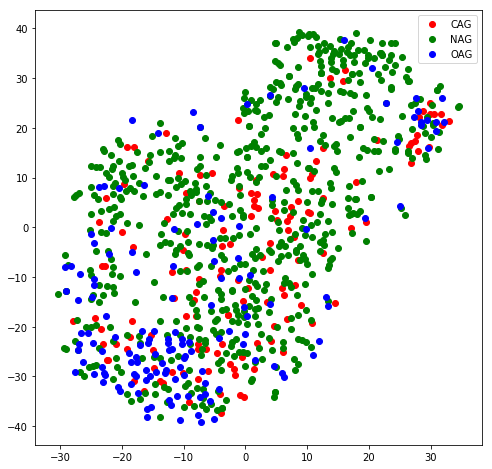}
\caption{Flatten vector of subnetwork1} 
\label{fig:sn1} 
\end{figure}
\begin{figure}[t!]
\centering
\includegraphics[width=.6\linewidth]{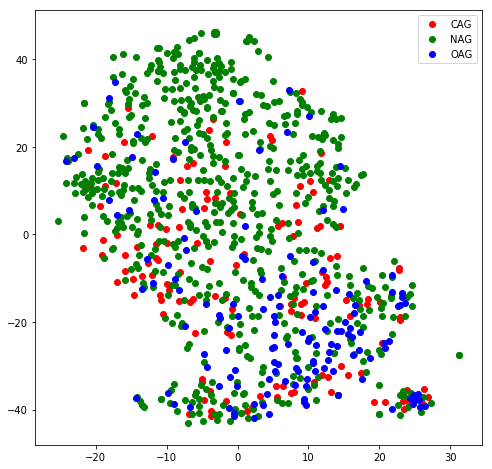}
\caption{Flatten vector of subnetwork2} 
\label{fig:sn2} 
\end{figure}
\begin{figure}[t!]
\centering
\includegraphics[width=.6\linewidth]{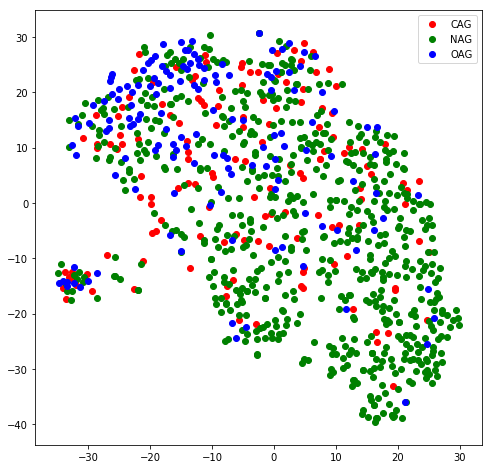}
\caption{Flatten vector of subnetwork3} 
\label{fig:sn3} 
\end{figure}

\begin{figure}[t!]
\centering
\includegraphics[width=.6\linewidth]{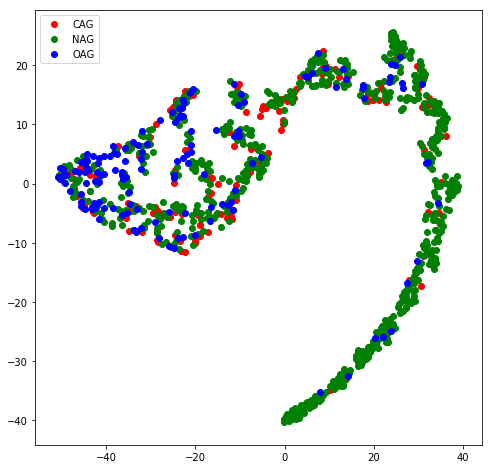}
\caption{Performance of Output Layer} 
\label{fig:dense} 
\end{figure}

\nocite{DBLP:journals/corr/abs-1804-00538}
Figures \ref{fig:sn1}, \ref{fig:sn2} and \ref{fig:sn3} are t-SNE \cite{vanDerMaaten2008} graphs, which depict the output of SN1-3 of CN1, respectively. We visualize the feature embeddings in all the SNs, and we observe that each SN is able to characterize the features distinctly due to the variability in the network configurations. When all the SNs are combined in an ensemble network, the feature representation is further improved. The inter-class variability is predominant, as can be validated in Fig. \ref{fig:dense}. This can be attributed to the fact that all 3 SNs have complimentary feature representations.

As observed from the confusion matrix of CN1 model ( Fig. \ref{fig:confusion matrix}), NAG is the easiest to detect. It is because most of the tweets in the data are NAG. The performance is better on OAG than on CAG, despite there being more training examples of CAG as OAG is more explicit and hence easier to identify, as opposed to the more indirect CAG \cite{hateoffensive}. CAG, because of its covert nature is the most difficult to classify. The confusion of CAG can also be observed in figure \ref{fig:dense}, where CAG is overlapping with NAG and OAG.

The confusion can also be seen by analyzing some \textbf{CAG} tweets \textbf{predicted} as \textbf{NAG}:\newline
"\textit{Hundreds of people were killed by your friends in Bombay, where were you at that time.}"\newline
"\textit{What's next? Soon we will be told to have a bullock cart and give up cars? Or live in a shed using candles?}" \newline
"\textit{Chit fund operators  n loan sharks r more honest}"

\section{Conclusion and Future Work}

We perform experiments to identify aggressive tweets by applying DL and Capsule Networks on preprocessed data. We show that capsulenets are efficient for aggression detection. We use an ensemble-based model and qualitatively show that each subnetwork learns unique features which help in classification. Our best model uses capsulenets and results in a 65.20\% f1 score, which is an improvement over most of the existing solutions.

In the future, we would like to explore other capsulenet architectures using different routing algorithms. A more in-depth analysis of CAG tweets could improve the performance on them.

\nocite{Wang:2018:SAC:3178876.3186015}
\nocite{safi-samghabadi-etal-2018-ritual}

\section{Acknowledgement}
We thank the anonymous reviewers for their constructive feedback. We also thank Rohan Sukumaran and Harshit Singh for fruitful discussions and proofreading. 
 \bibliographystyle{acl_natbib}
\bibliography{references}
\end{document}